\documentclass{llncs}

\usepackage{amsmath,amssymb,stmaryrd,url}
\usepackage{graphicx}
\usepackage{multirow}
\usepackage[utf8]{inputenc}
\usepackage[english]{babel}
\usepackage{xspace}
\usepackage[usenames, dvipsnames]{color}
\usepackage{wrapfig}

\newcommand{\prob}[1]{\textit{avatar aliases identification problem}}
\newcommand{\starcraft}[1]{\textit{Starcraft~2}}
\title{Identifying Avatar Aliases in Starcraft~2\thanks{This research has been partially funded by the French National Project FUI AAP 14 Tracaverre 2012-2016.}}
\titlerunning{Identifying Avatar Aliases in Starcraft~2}  
\author{Olivier Cavadenti, Victor Codocedo, Jean-François Boulicaut, Mehdi Kaytoue}
\authorrunning{Cavadenti et al.} 
\institute{
Universit\'e de Lyon. CNRS, INSA-Lyon, LIRIS. UMR5205, F-69621, France.\\
contact:\textit{firstname.name@insa-lyon.fr}
}
\begin{document}
\maketitle

%----------------ABSTRACT-------------------%
\begin{abstract}
In electronic sports, cyberathletes conceal their online training using
different avatars (virtual identities),
allowing them not being recognized by the opponents
they may face in future competitions. In this article,
we propose a method to tackle this \prob{}.
Our method trains a classifier on behavioural data
and processes the confusion matrix to output label pairs
which concentrate confusion. 
We experimented  with \starcraft{} and report our first results.
\end{abstract}

%----------------INTRODUCTION-------------------%
\section{Introduction}
 
In most of online competitive games, 
players need an ``avatar'' (an online identity) to log in the game network.
Nothing forbids a player to have several avatars
and actually, it is a very common practice for cyberathletes.
Players generally have one official avatar for official tournaments,
and several others to conceal their game tactics without being recognized by other players
they may meet online: global rankings and leagues are public just as in chess and tennis,
while game logs are available and prone to analysis by means of visualization and
machine learning just as in standard sport analytics.
Accordingly, we are facing a set of players, generating behavioural data,
in an unknown one-to-many relationship with avatars (handling \textit{many-to-many}
relationships is left to future work).
In this context, the \prob{} aims at discovering the group of avatars belonging to the same player.
Solving this problem is motivated by the growing need of e-sport structures to
study the games and strategies of the opponents (match preparation),
and the security challenges of game editors (detecting avatar usurpers).

Yan et al. showed that a classifier can be trained to predict 
with high accuracy the avatars involved in a game play 
of \starcraft{}~\cite{DBLP:conf/chi/YanHC15}.
Nevertheless, they purposely considered datasets without players having several avatars 
(what we call avatar aliases): in presence of such \textit{avatar aliases},
the prediction accuracy drastically degrades,
since prediction models fail at differentiating two avatars of the same player.
We extend this work and answer the  \prob{}:
it relies on mining the confusion matrix yielded by a supervised classifier
using Formal Concept Analysis~\cite{ganter99},
and exploits the confusion a classifier has in presence of avatar aliases
 when they belong to the same player.
Experimental evaluation shows promising results.

%----------------MODEL-------------------%
\section{Basic notations and general intuition\label{sec:method}}
\newcommand{\nconf}{\ensuremath{\tilde{C}^\rho}}
\newcommand{\conf}{\ensuremath{C^\rho}}

Let $A$ be a set of avatars and $T$ be a set of traces such as for a given avatar $a \in A$, the set $T_a \subseteq T$ is the set of all traces generated by $a$.  Consider a classifier $\rho$ where labels are the avatars to predict. A classifier is a function $\rho: T \rightarrow A$ that assigns the avatar $\rho(t) \in A$ to a given trace $t \in T$. 
Let $n = |A|$ be the number of avatars in $A$,
from any classifier $\rho$, one can derive a confusion matrix %$\conf_{|A \times A|} = [c_{ij}]$
$\conf_{n \times n} = (c_{i,j})$ %_{1 \leq i \leq  \vert A \vert, 1 \leq j \leq  \vert A \vert} $$
where $c_{i,j} =  \vert \{ t \in T_{a_i} ~s.t.~ \rho(t) = a_j \} \vert$.
Each row and column of $\conf$ correspond to an avatar, 
while the value $c_{ij}$ is the number of traces of avatar $a_i$
that are classified by $\rho$ as of avatar $a_j$.
The normalized confusion matrix is given by
$\nconf = [c_{i,j}/|T_{a_i}|]$
where $\nconf_{i,i} = 1$ for any $i \in [1, |A|]$ means all the traces of avatar $a_i$ are correctly classified by $\rho$. 

\begin{wrapfigure}{r}{40mm}\centering
{\scriptsize
\vspace{-7mm}
\begin{tabular}{|c|c|c|c|c|c|}
\hline
& $a_1$ & $a_2$ & $a_3$ & $a_4$ & $a_5$ \\  
\hline
$a_1$ & 0.6 & 0.4 & 0 & 0 & 0 \\ \hline
$a_2$ & 0.4 & 0.55 & 0.05 & 0 & 0 \\ \hline
$a_3$ & 0 & 0 & 0.8 & 0.15 & 0.05 \\ \hline
$a_4$ & 0 & 0.05 & 0 & 0.7 & 0.25 \\ \hline
$a_5$ & 0 & 0 & 0 & 0.5 & 0.5 \\ \hline
\end{tabular}
\caption{\scriptsize Confusion matrix}
\label{table.1}
}

\vspace{-8mm}
\end{wrapfigure}
Our goal is to discover the group of avatars that belong to the same player.
Our intuition is that a classifier will hardly differentiate these avatar aliases, 
hence the confusion matrix values should be high and concentrated around them.
This is exemplified in Figure \ref{table.1}: avatars $\{a_1, a_2\}$ are candidates 
to belong to the same player, $\{a_4, a_5\}$ shall belong to another player,
while ${a_3}$ stays as singleton with a diagonal high value. 
A reasonable clustering of avatars would be given by 
$\{ \{a_1, a_2\}, \{ a_3 \}, \{a_4, a_5\}\}$.

More formally, given a normalized confusion matrix $\nconf$, 
we would like to find pairs of avatars $a_i,a_j \in U$ 
such that $\nconf_{ij} \simeq \nconf_{ji} \simeq \nconf_{ii} \simeq \nconf_{jj}$
and $\nconf_{ij} + \nconf_{ji} + \nconf_{ii} + \nconf_{jj} \simeq 2$.
These conditions come from the fact that, if $a_i,a_j$ correspond to the same player, 
traces in $T_{a_i}$ have the same probability of being classified as $a_i$ or $a_j$ 
(the same for traces in $T_{a_j}$). 
Furthermore, for a trace of avatar $a_i$, it is required that the probability of classification is spread between $a_i$ and $a_j$ only,
meaning that $\nconf_{ij} + \nconf_{ii} \simeq 1$ (similarly for $a_j$).

%Using this rationale, in what follows we propose (i) to extract patterns from the confusion matrix,
%and (ii) to post process them to provide groups of candidate avatar pairs.
%The first step is achieved thanks to Formal Concept analysis (FCA \cite{ganter99,ganter01a}),
%while we define scoring functions and ranking for the second step.

\section{Method}
Our method firstly extracts fuzzy concepts from 
the confusion matrix, scores and post-processes
them to generate avatar pairs, 
candidates to be aliases.

\medskip

\noindent\textbf{Fuzzy concepts in a confusion matrix.} Let us define the fuzzy set of membership degrees $L^A$ where $L = [0,1]$, 
such as the mapping function $\delta:A \rightarrow L^A$ assigns membership values 
for the avatar $a_i$ in the fuzzy set $L^A$ based on the normalized confusion matrix. 
Simply, this is a mapping that assigns to $a_i$ its corresponding row in $\nconf$ which we denote $\nconf_i$.  We model a confusion matrix $\nconf$ as a pattern structure $(A,(L^A,\sqcap),\delta)$ \cite{ganter01a}.
The operator $\sqcap$ is a meet operator in a semi-lattice (idempotent, commutative and associative),
and is defined as follows, given two avatars $a_i,a_j \in A$:
\begin{align*}
\delta(a_i) \sqcap \delta(a_j) &= \langle min(\nconf_{ik}, \nconf_{jk})  \rangle,\, k \in [1,|A|] \\
\delta(a_i) \sqsubseteq \delta(a_j) &\iff \delta(a_i) \sqcap \delta(a_j) = \delta(a_i)
\end{align*}

\smallskip

\noindent\textit{Example}. The Figure \ref{table.1} illustrates a confusion matrix
obtained from a classifier $\rho$. We have 
$\delta(a_1) = \{ a_1^{0.6}, a_2^{0.4}, a_3^{0}, a_4^0, a_5^0 \}$,
$\delta(a_2) = \{ a_1^{0.4}, a_2^{0.55}, a_3^{0.05}, a_4^0, a_5^0 \}$
and $\delta(a_1) \sqcap \delta(a_2) = \{ a_1^{0.4}, a_2^{0.4}, a_3^{0}, a_4^0, a_5^0 \} $.
\smallskip

\newcommand{\extent}{\ensuremath{\mathtt{A}}}
\newcommand{\pattern}{\ensuremath{\mathtt{d}}}
\newcommand{\pc}[1][]{\ensuremath{\mathtt{(A_{#1},d_{#1})}}}
Actually, $\sqcap$ corresponds to the fuzzy set intersection and $(L^A,\sqsubseteq)$ 
is a partial order over the elements of $L^A$ which can be represented as a semi-lattice. %
The pattern structure  $(A,(L^A,\sqcap),\delta)$ is provided with two derivation operators,
forming a Galois connection \cite{ganter01a}. Formally, we have, for a subset of avatars $\extent \subseteq A$ 
and a fuzzy set $\pattern \in L^A$ such as: $\extent^\square = \bigsqcap_{a \in \extent} \delta(a)$ and $\pattern^\square = \{a \in A ~|~ \pattern \sqsubseteq \delta(a) \}$.
The pair $\pc$ is a pattern concept iff $\extent^\square = \pattern$ and $\pattern^\square = \extent$. Pattern concepts are ordered by extent inclusion such that for $\pc[1]$ and $\pc[2]$ we have:
$\pc[1] \leq \pc[2] \iff \extent_1 \subseteq \extent_2 \, (\text{or }\pattern_1 \sqsupseteq \pattern_2)$. A pattern concept $\pc$ contains a fuzzy set $\pattern$ which can be represented as a \emph{vector} $\pattern = \langle \pattern^j \rangle$ with length $|A|$ where each value $\pattern^j$ 
is the minimum for all rows $i$ in column $j$ of matrix $\nconf$ s.t. $a_i \in \extent$.  

\medskip

\noindent\textbf{Computing and scoring concepts.} %
From the confusion matrix we compute all possible pattern concepts using the \texttt{addIntent} algorithm~\cite{vandermerwe2004addintent}. Pattern concepts are then ranked according to a score and converted into a list of pairs. For example, if a pattern concept extent contains three avatars $a_1,a_2$ and $a_3$, we convert this concept into pairs $(a_1,a_2)$,  $(a_1,a_3)$ and $(a_2,a_3)$. The scoring function  $s:L^A \rightarrow [0,1]$ is given as follows: for a pattern $\pattern$, $s(\pattern) = \Sigma_{j = 1}^{|A|} \pattern^j $. 

\smallskip

\noindent\textit{Example}. In Figure \ref{table.1}, we have:
$s(\{a_1,a_2\}^\square) = 0.8$, $s(\{a_4,a_5\}^\square) = 0.75$ and
$s(\{a_1,a_2,a_4\}^\square) = 0.05$.

\smallskip

It is clear that the function $s$ is decreasing w.r.t. the order of pattern concepts, i.e. $\pc[1] \leq \pc[2]  \implies s(\pattern_1) \leq s(\pattern_2)$. 
Thus, pattern concepts can be mined up to a given score threshold analogously
as formal concepts can be mined up to a given minimal support.
We can appreciate that the higher the score of a given pattern, 
the more \emph{confused} is the classification of traces of avatars $a \in \extent$ by $\rho$ in $\nconf$ and thus,
they become candidates for merging. 
This property directly follows from the choice of our similarity operator $\sqcap$ as a fuzzy set intersection,
which behaves as a pessimistic operator (returning minimum values).

\medskip

\noindent\textbf{Ranking avatar aliases.} %
Consider the clustering condition previously formalized as $\nconf_{ij} \simeq \nconf_{ji} \simeq \nconf_{ii} \simeq \nconf_{jj}$ and $\nconf_{ii} + \nconf_{ij} + \nconf_{ji} + \nconf_{jj} \simeq 2$. Consider that the pair of avatars ($a_i,a_j$) respects these conditions. It is easy to see that ($a_i,a_j$) will necessarily be a candidate pair highly ranked from the previous step. 
\begin{align*}
\nconf_{ij} \simeq \nconf_{jj} \simeq min(\nconf_{ij},\nconf_{jj}) 
\mathrm{~~and~~}
\nconf_{ii} \simeq \nconf_{ji} \simeq min(\nconf_{ii},\nconf_{ji}) \\
\implies min(\nconf_{ij},\nconf_{jj}) + min(\nconf_{ii},\nconf_{ji}) \simeq 1
\end{align*}
Thus, the set of avatar clusters we are looking for are contained within the set of candidate pairs and moreover, they are highly ranked. In order to remove pairs from the list of candidates that do not hold the avatar cluster definition, we propose a cosine similarity measure between a couple of vectors calculated for each avatar as follows. Let ($a_i,a_j$) be a candidate pair, the cluster score is defined as: $cluster\_score(a_i,a_j) = cosine(\langle \nconf_{ii}, \nconf_{ij}\rangle, \langle \nconf_{jj}, \nconf_{ji}\rangle)$.

\begin{wrapfigure}{r}{15mm}\centering
{
%	\vspace{-8mm}
\scriptsize
	\begin{tabular}{|c|c|c|}
		\hline
		& $a_i$ & $a_j$ \\ 		\hline
		$a_i$ & 1 & 0 \\		\hline
		$a_j$ & 1 & 0 \\ 		\hline
	\end{tabular}}
	\vspace{-8mm}
\end{wrapfigure}
The cluster score establishes a measure of how close is a candidate pair from being an avatar cluster.
The logic comes from the following scenario. Consider that the traces of avatar $a_i$ were all correctly classified meaning that $\nconf_{ii} = 1$ and that the traces of avatar $a_j$ were all incorrectly classified as $a_i$, meaning that $\nconf_{ji} = 1$, thus we have the section of the normalized confusion matrix illustrated on the right hand side. We can observe that the pair $(a_i,a_j)$ will be contained in the set of candidate pairs and will be highly ranked, even though it is not an avatar cluster since it violates the first condition. The cluster score for this particular case can be calculated as:
$cluster\_score(a_i,a_j) = cosine (\langle 1,0 \rangle, \langle 0,1 \rangle) = 0$, 
meaning that this candidate pair is not an avatar cluster. Notice that for the pair of avatars such that $a_{ii} = 1$ and $a_{jj} = 1$, the cluster score is 1 (cosine between parallel vectors) while the pair is not an avatar cluster. However this pair would have a score $s$ equal to 0 and would be at the bottom of the ranked candidate pairs. A third kind of pair occurs when the traces of $a_i$ and $a_j$ are all incorrectly classified as a third avatar $a_k$. In such a case, the cluster score is 0.
The post processing step is executed as follows. Given a ranked list of candidate pairs yielded from the previous step, each pair is evaluated using the cluster score. Given an arbitrary threshold $\lambda$, if the cluster score of the candidate pair is below this threshold, then it is rejected. Candidate pairs are re-ranked into a final list of avatar clusters.

%----------------EXPERIMENTS-------------------%

\section{Experiments}
\label{sec:xp}

\noindent\textbf{Data collections and objectives.} 
We constructed two collections of Starcraft 2 replays to test our method. A replay contains all
data necessary for the game engine to replay the game. Replays are shared on dedicated 
websites\footnote{\url{http://wiki.teamliquid.net/starcraft2/Replay_Websites}} and can be parsed to extract relevant features\footnote{\url{http://sc2reader.readthedocs.org/}}. The first collection has been chosen for studying the accuracy of classifiers to recognize avatars from their traces:
we have selected 955 professionals games of 171 unique players which cannot contain avatar aliases\footnote{\url{http://wcs.battle.net/sc2/en/articles/wcs-2014-season-2-replays}}. The second collection, which have possible avatar aliases, is built with all replays available on \textit{SpawningTool.com} in July 2014, for a total 10,108 one-versus-one games and 3,805 players. This collection corresponds to a real world situation, and is used for evaluating our avatar alias resolution approach.

\medskip

\begin{figure}[h!]
	\centering
	\vspace{-0.6cm}
		\hfill
		\includegraphics[width=0.51\textwidth]{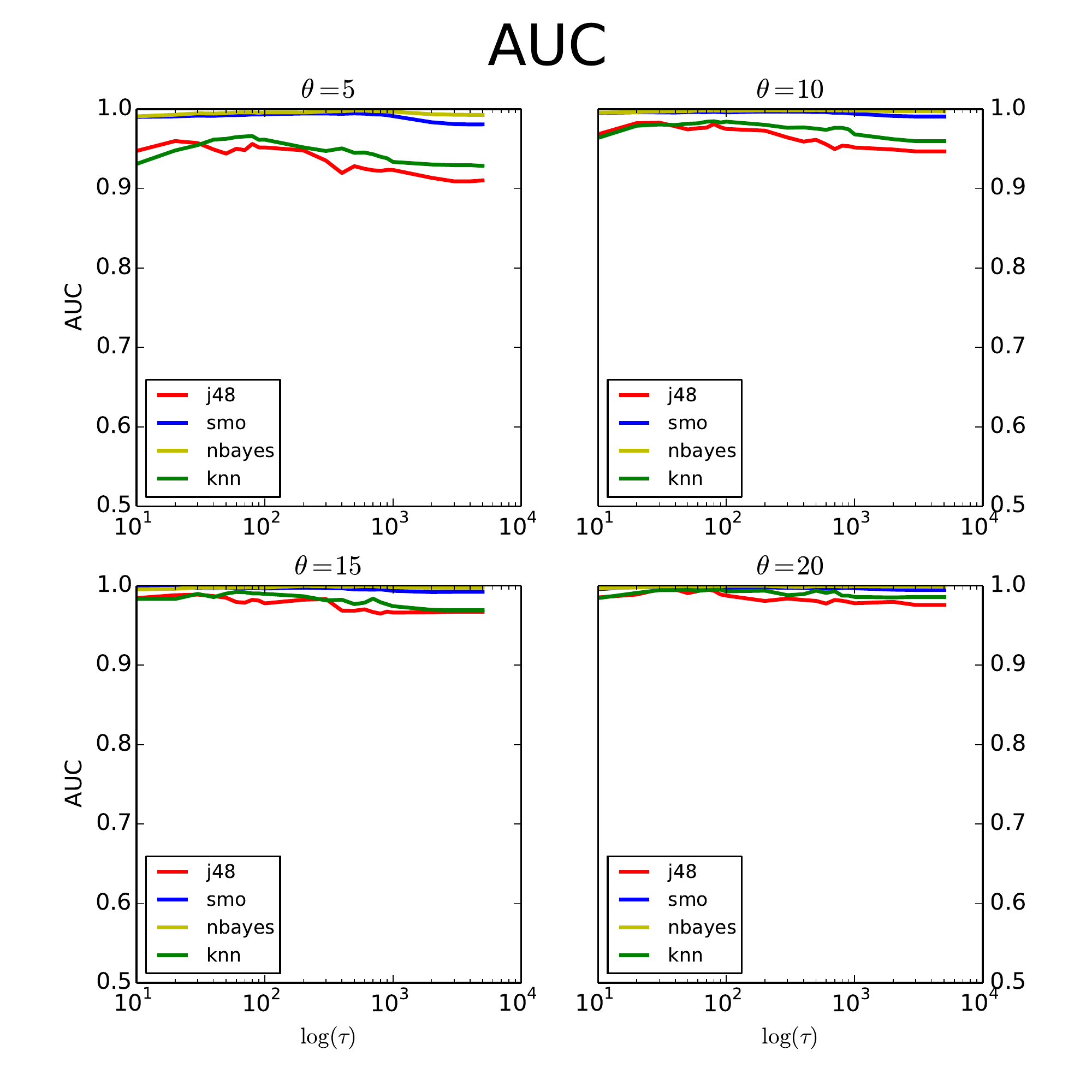}
	\hspace{-0.5cm}
		\includegraphics[width=0.51\textwidth]{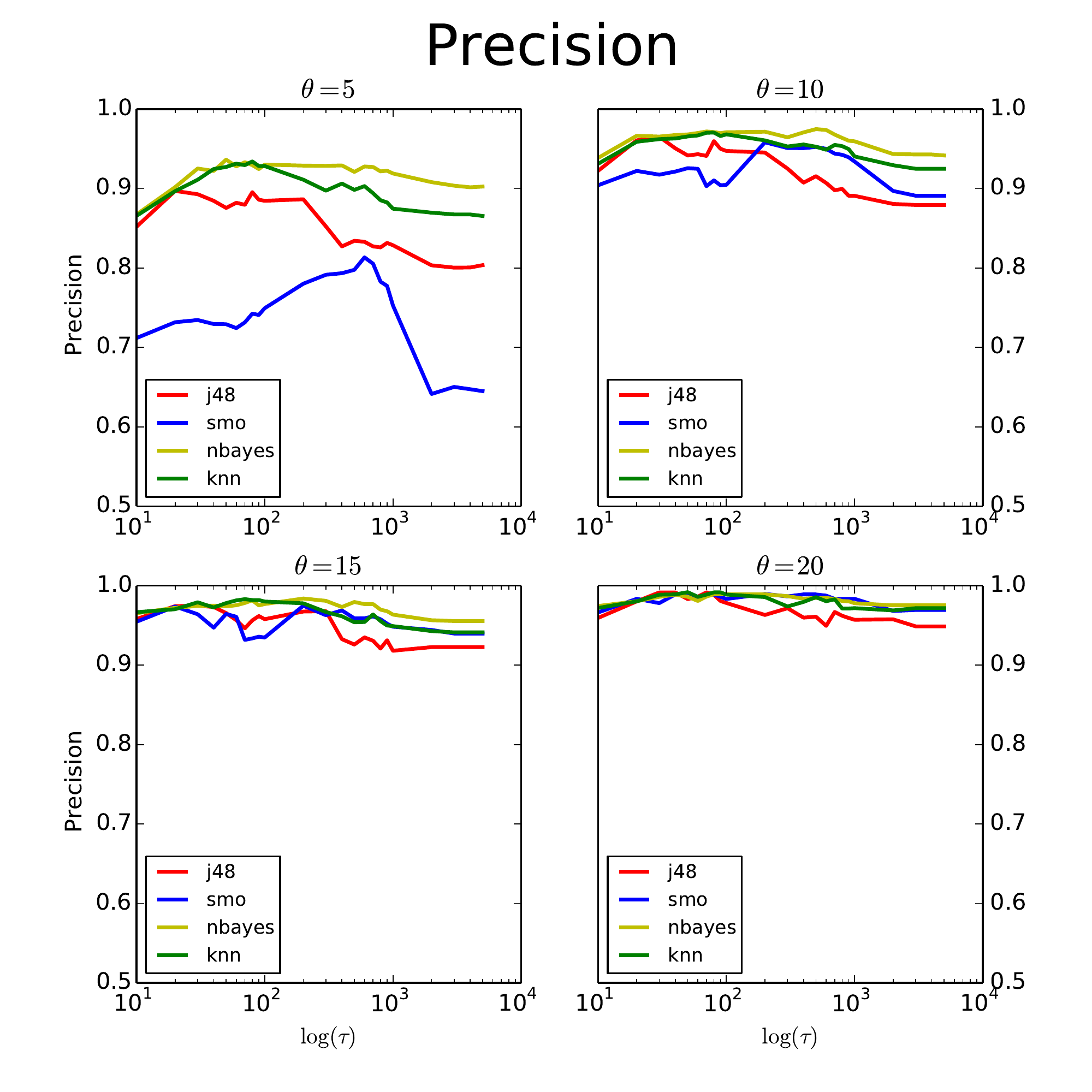}
		\hfill
\vspace{-0.6cm}
	\caption{\scriptsize Classification results for Collection 1: ROC area under the curve (AUC) and precision distribution on 23 points of $\tau$ for four $\theta$ values.
	}
	\label{fig:results.1}
\vspace{-0.5cm}
\end{figure}

\noindent\textbf{Classifying avatars.}
Our method analyses the confusion matrix of a given classifier $\rho$.
Good features, as well as a prediction method, should first be chosen.
As features, we use the hotkey usage count \cite{DBLP:conf/chi/YanHC15}
during the $\tau$ first seconds of the game:
there are 30 of such features ($\{0,...,9\} \times \{assign, remove, select\})$.
We also consider the \textit{faction} of the player,
the game outcome (\textit{winner} or \textit{loser}) and actions per minutes in average (APM).
We generated several datasets given the $\tau$ parameter, 
and introduced also a minimum number $\theta$ of games an avatar should have to be considered in the dataset.
Each dataset is classified using the Weka machine learning software \footnote{\url{http://www.cs.waikato.ac.nz/ml/weka/}} and evaluated using 10-fold cross validation from which we obtain a confusion matrix. We chose four different classifiers, namely K Nearest Neighbours (knn), Naive Bayes (nbayes), J48 decision tree (j48) and Sequential Minimization Optimization (smo). Parameters for each of the classifier were left as default. Figure \ref{fig:results.1} shows the ROC area and the precision obtained for 92 datasets created for Collection 1. The parameter $\tau$ ranged over 23 values in an exponential scale, initially from 10 to 90 seconds then from 100 to 900 and finally from 1000 to 5000 seconds (the longest game in this collection has around 5300 seconds) and thus, the x axis of each figure is in logarithmic scale. 
For each measure, four figures corresponding to four different settings of $\theta$ are presented.
Each line corresponds to a different classifier. The figures present an empirical evaluation that the initial assumption, that avatars are very easily recognizable based in the signatures left in the traces they generate while playing, is true. For each different setting, ROC area is always around 100\% showing the robustness of the approach under different parametrizations. Precision is always maintained over 60\%, achieving its minimal value for the SMO classifier with $\theta = 5$ and $\tau > 1000$.  Actually, this also supports the following assumptions. Firstly, it is hard to recognize users that have played a few games, meaning that the larger the value of the $\theta$ threshold, the more discriminative power has the classifier. Secondly, users are recognizable in the first few minutes of the game. The precision curves show a slight concave behaviour hinting a maximum of the precision w.r.t. the time cut used for traces. Users can be efficiently discovered by their hotkeys binding settings. As the game progresses, traces may differ given that the number of options in the game greatly increase and vary in execution regarding different opponents.

\medskip

\noindent\textbf{Main method evaluation strategy.}
As we do not have information about the users behind the avatars,
it is not possible to evaluate the avatar pairs candidates using a ``ground truth''.
Hence we performed an evaluation of our approach using three different strategies.
First consider that an avatar of \starcraft2{} is given by its 
\textit{Battle.net account URL}, made of a server name (Europe, America, etc.),
a unique identifier, and an avatar name. We use the whole URL as avatar class labels
in our classifiers. 
Note now that players have several accounts, on different servers, that may share the name.
Players can also change the name of their avatar: 
it does not affect the ID and server that identify their account.
As our method returns an ordered list of pairs candidates merging,
we consider the following indicators, for each pair.
\begin{itemize}
\item \textit{Avatar names.} Two avatars may have the same name but different battle net id. It is weak indicator as it can be a common name (e.g. \textit{Batman)}.
\item \textit{Battle.net account unique ID.} Two avatars may have two different names but the same unique identifier. This is a strong indicator.
\item \textit{Surrogates.} We create surrogate avatars $a^1,a^2$ from an avatar $a \in A$ by generating a partition in two different subsets of traces for each avatars. Our goal is to retrieve that $a^1$ and $a^2$ are avatar aliases. For splitting traces of an avatar into surrogates, 
we introduce a parameter $\beta$ as a balance between the traces distributed over the surrogate avatars
($\beta = 0.5$ yields that both surrogate avatars will have half associated traces). 
We introduce others parameters:  $\gamma$ is the proportion of avatars who are converted in surrogates aliases. We assume that professionals play a lot of games then we select the avatar which have played more than $\theta$ games. As we have observed, it is not necessary to analyse the entire replay to discriminate an avatar then we select the first $\tau$ either actions or seconds. 
\end{itemize}

\begin{wrapfigure}{r}{5.5cm}\centering
	\vspace{-1.2cm}
	\includegraphics[width=1\linewidth]{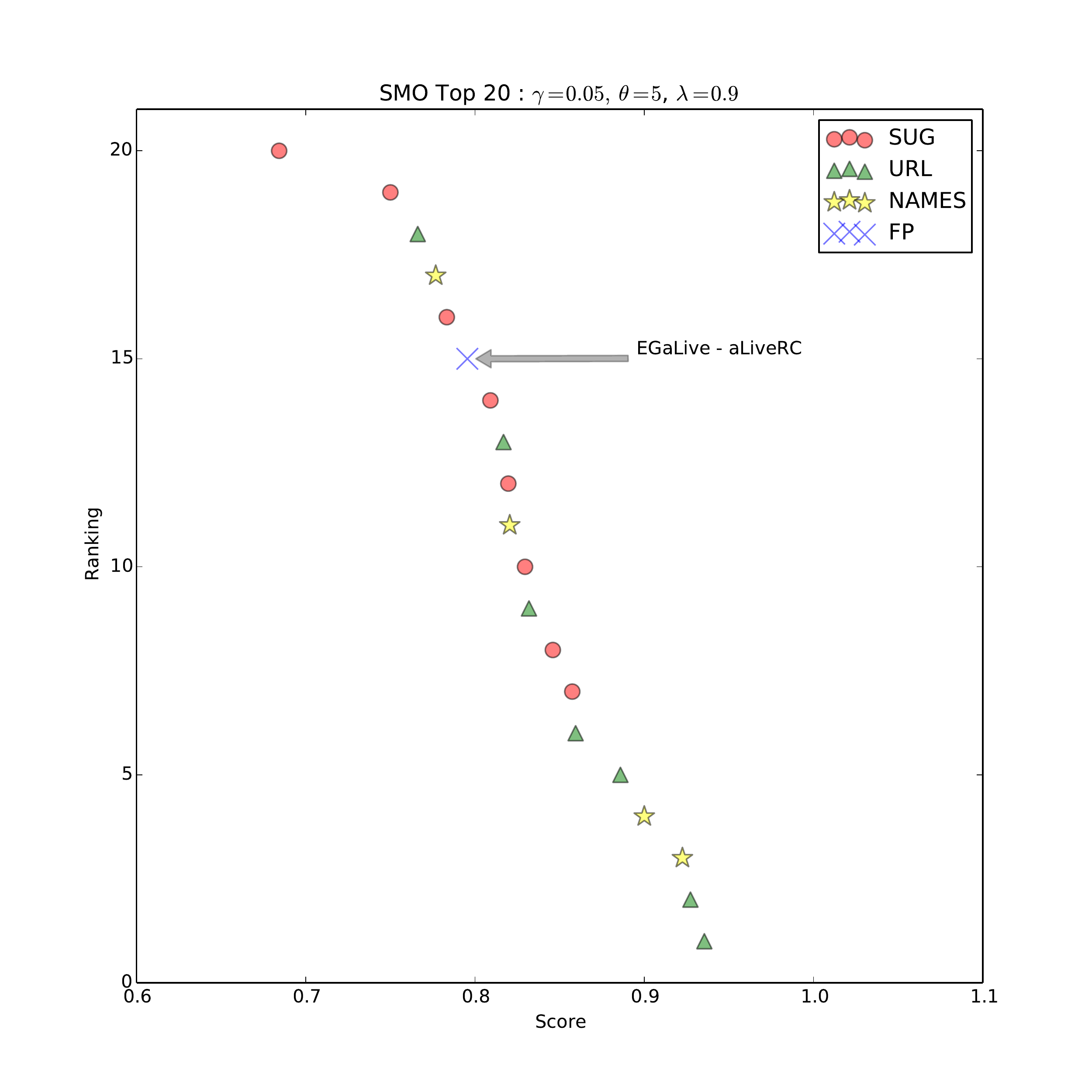}
	\vspace{-1.2cm}
	\caption{Candidate pairs ranking\label{fig:results.lambda}.}
	\vspace{-0.7cm}
\end{wrapfigure}

To evaluate our approach we will measure the precision, recall and f-measure of the first 100 ranked  avatar clusters. Given the ranking $r$, we consider $TP,FP$ and $FN$ stand for true positives, false positives and false negatives respectively. A pair is a false positive when we do not have enough information to consider them as true positives, meaning that their avatar names do not match, their URL is different and they are not part of our own set of surrogate avatars. \emph{They are in fact the kind of pairs we are looking for}.
As an example, the Figure~\ref{fig:results.lambda} shows the initial candidate pairs extracted from a confusion matrix generated by a Sequential Minimization Optimization (SMO) classification with $\gamma = 0.05$, $\theta = 5$ and $\lambda~=~0.9$. Within the figure, a point represent a pair of avatars with a red circle if the avatars are surrogates, a green triangle if they have the same account, a yellow star if they have the same name and in the other cases a blue cross, annotated with the nick-names of the avatars. The only FP in this list is a couple of avatars that belong to the player known as \texttt{aLive}\footnote{\url{http://wiki.teamliquid.net/starcraft2/ALive}}.  We also report on other three measures, namely P@10 (precision in the first 10 elements of the ranking), mean average precision (MAP), the receiver operating characteristic (ROC) and the ROC area under the curve (AUC).

%----------------RESULTS-------------------%

\medskip

\noindent\textbf{Identifying multiple aliases.} The goal of these experiments is to assess our approach for finding avatar aliases based of the evaluation introduced above. For generating datasets, we have selected three different $\tau$ values, namely $30, 60$ and $90$ seconds. We have picked the same values for $\theta$ as in the previous experiments. Surrogates were generated for the first $5$, $10$, $15$ and $20$ percent of the most active users in the dataset ($\gamma$) and we have set the balance $\beta = 0.5$. For each of the previously selected classifiers, the confusion matrix was processed by the Sephirot addIntent implementation\footnote{https://code.google.com/p/sephirot/} to obtain a set of pattern concepts. Scoring and post processing were implemented in ad-hoc python scripts. 
Table~\ref{table.avatars.details.1} shows a summary for the evaluation results using the top 100 pairs of avatar clusters. Results indicate that our approach is very efficient at identifying surrogate avatars, particularly for KNN and the J48 classifiers achieving very high recall values. In the upper part of the table, while precision is low it is worth noticing that in the top 100, there are only 41 surrogates meaning that the maximum achievable precision is 0.41. The classifier KNN is particularly good in this measure achieving an almost perfect value (0.4 of 0.41). All four classifiers achieve a very high precision in the first 10 results (P@10) while two of them get a perfect score. Indeed, one of the main characteristics of our approach is the good ranking it generates over the avatar pairs. This fact is confirmed by the good MAP and ROC area under the curve (AUC) values achieved by all four classifiers. Both these measures slightly degrade when including in the set of true positives URLs and names. This can be understood since not all avatars with the same name necessarily belong to the same user. Thus, pairs of avatars with the same name will be more evenly distributed over the ranking or can even be found at the bottom indicating that they do not belong to the same user. This fact is reflected in the gap between the high growth of precision and low degradation of recall, i.e. avatars with the same name are distributed between the pairs retrieved and those that were not. As we have discussed, avatars with the same URL necessarily belong to the same user. Hence, we would have expected that in the first 10 pairs retrieved we could find an even distribution of surrogates and URLs. Instead, for all classifiers, P@10 is more than 80\% surrogates (while the rest is always URLs - P@10 in the medium part of the table).  Table~\ref{table.avatars.details.2} shows a summary of results when looking for just surrogates while varying the balance in the distribution of traces between them. We can clearly observe that the performance of the approach quickly degrades as more imbalanced gets the distribution (the higher the $\beta$ value). Actually, for some classifiers it is not possible to obtain a single good result, even when we have lowered the $\lambda$ threshold to 0.8. As URLs are not necessarily balanced, classifiers tend to predict the label of a trace belonging to an avatar with less traces to one with more traces. Issues related to learning from imbalanced datasets are reviewed in~\cite{He:2009:LID:1591901.1592322} and need to be considered when selecting a proper classifier for our particular application.

\begin{table*}[ht!]
	\begin{minipage}[c]{.50\linewidth}
	\scriptsize
	\centering
{	\begin{tabular}{lcccccc}
		\hline
		\multicolumn{7}{c}{\textbf{Parameters:$\,\gamma = 0.2,\,\theta = 20,\,\lambda = 0.9,\, \tau = 90$}} \\
		\hline\hline
\multicolumn{7}{l}{\textbf{SUG}} \\
Cl & F1 & MAP & Recall & AUC & Prec. & P@10\\ 
\hline
$j48$&0.468 & 0.824& 0.805 & 0.904 & 0.33 & 1.0\\ 
$nbayes$&0.226& 0.740 & 0.390  & 0.915  & 0.16 & 0.8\\ 
$smo$& 0.312 & 0.971 & 0.536 & 0.993  & 0.22 & 1.0\\ 
$knn$&0.567& 0.822 & 0.976& 0.882 & 0.4 & 0.9\\ 
\hline\hline
\multicolumn{7}{l}{\textbf{SUG \& URLS}} \\
Cl &F1 & MAP & Recall & AUC & Prec. & P@10\\ 
\hline
$j48$&0.588& 0.907 & 0.606 & 0.866 & 0.57 & 1.0\\ 
$nbayes$&0.443 & 0.857 & 0.457 & 0.864 & 0.43 & 1.0\\ 
$smo$&0.257 & 0.912 & 0.266 & 0.945 & 0.25 & 1.0\\ 
$knn$&0.670 & 0.937 & 0.691 & 0.874 & 0.65 & 1.0\\ 
\hline\hline
\multicolumn{7}{l}{\textbf{SUG \& URLS \& Names}} \\
Cl&F1 & MAP & Recall & AUC & Prec.& P@10\\ 
\hline
$j48$&0.689 & 0.983 & 0.606 & 0.935 & 0.8 & 1.0\\ 
$nbayes$&0.560 & 0.943 & 0.492 & 0.906 & 0.65 & 1.0\\ 
$smo$&0.258 & 0.949 & 0.227 & 0.960 & 0.3 & 1.0\\ 
$knn$&0.758 & 0.967 & 0.667 & 0.792 & 0.88 & 1.0\\ 
	\end{tabular}
	\caption{Main results}
	\label{table.avatars.details.1}}
\end{minipage}
\hfill
\begin{minipage}[c]{.50\linewidth}
\scriptsize
		\centering{
	\begin{tabular}{lcccccc}
		\hline
		\multicolumn{7}{c}{\textbf{Parameters:$\,\gamma = 0.2,\,\theta = 20,\,\lambda = 0.8,\, \tau = 90$}} \\
		\hline\hline
		\multicolumn{7}{l}{\textbf{J48}} \\
		Balance &F1 & MAP & Recall & AUC & Prec. & P@10\\ 
		\hline
$\beta = 0.5$&0.925 & 0.996 & 0.929 & 0.955 & 0.920 & 1.0\\ 
$\beta = 0.6$&0.545 & 0.927 & 0.632 & 0.921 & 0.480 & 1.0\\ 
$\beta = 0.7$&0.053 & 0.695 & 0.077 & 0.977 & 0.040 & 0.3\\ 

		\hline\hline
		\multicolumn{7}{l}{\textbf{Naive Bayes}} \\
		Balance &F1 & MAP & Recall & AUC & Prec. & P@10\\ 
		\hline
$\beta = 0.5$&0.472 & 0.902 & 0.475 & 0.953 & 0.470 & 0.9\\ 
$\beta = 0.6$&0.273 & 0.923 & 0.316 & 0.973 & 0.240 & 1.0\\ 
$\beta = 0.7$&0.197 & 0.9 & 0.288 & 0.978 & 0.150 & 0.9\\ 
$\beta = 0.8$&0.048 & 0.533 & 0.120 & 0.983 & 0.030 & 0.3\\ 
		\hline\hline
		\multicolumn{7}{l}{\textbf{SMO}} \\
		Balance&F1 & MAP & Recall & AUC & Prec. & P@10\\ 
		\hline
$\beta = 0.5$&0.392 & 0.983 & 0.394 & 0.992 & 0.390 & 1.0\\ 

		\hline\hline
		\multicolumn{7}{l}{\textbf{KNN}} \\
		Balance&F1 & MAP & Recall & AUC & Prec. & P@10\\ 
		\hline
$\beta = 0.5$&0.905 & 0.964 & 0.909 & 0.732 & 0.9 & 1.0\\ 
$\beta = 0.6$&0.750 & 0.957 & 0.868 & 0.929 & 0.660 & 1.0\\ 
$\beta = 0.7$&0.184 & 0.706 & 0.269 & 0.949 & 0.140 & 0.7\\ 
%$\beta = 0.8$&- & - & - & - & - & -\\ 
	\end{tabular}
	\caption{Varying surrogate balance ($\beta$)}
	\label{table.avatars.details.2}}
\end{minipage}
\end{table*}

%----------------CONCLUSION-------------------%
\vspace{-1.3cm}
\section{Conclusion}
\label{sect:conclusions}
We introduced the problem of avatar aliases identification
when there exists no mapping between individuals and their avatars.
This is an important problem for game editors, but also for e-sport structures.
Our method relies on the fact that behavioural data hide individual characteristic patterns,
which allows making predictive approaches very accurately. 
Nevertheless, this good performance quickly degrades when data hides avatar aliases, 
which is why we based our analysis on confusion matrices.
As future work, we plan to study other competitive games, 
and how biclustering could tackle the problem.
We also believe that our approach can be used to solve other application problems,
such as identifying users on different devices (smart-phones, tablet, computer, etc.)
regarding the usage traces they left.

\bibliographystyle{plain}
\bibliography{MLSA2015}
\end{document}